\RequirePackage[T1]{fontenc}
\documentclass[letterpaper,10pt,conference]{ieeeconf}
\IEEEoverridecommandlockouts
\usepackage{amsmath,amssymb}
\DeclareMathSizes{10}{10}{7}{7}
\usepackage{graphicx}
\usepackage{cite}
\usepackage{url}
\usepackage{balance}
\usepackage[draft]{hyperref}
\graphicspath{{figures/}}
\title{\LARGE\bf CereVLA: Cerebellum-Inspired Consequence-Aware Residual Governance for Efficient Vision-Language-Action Execution}
\author{Shuai Zeng, Yuxuan Liang, Hangmiao Hu, Fobao Zhou,
Zixiang Wang, Wenxi Hong, and Hang Zhao%
\thanks{The authors are with the Hong Kong University of Science and Technology (Guangzhou), Guangzhou, China (e-mail: hangzhao@hkust-gz.edu.cn). Hang Zhao is the corresponding author.}%
}
\newif\iffirstpageteaser
\firstpageteasertrue
\newcommand{\overviewlabel}{\label{fig:overview}}
\newcommand{\overviewcaption}{Overview of CereVLA. FRR samples corrections,
selects a candidate using an expert-supervised reference head, and applies
learned scalar modulation. One-step evaluation ($H_1$) and history-aware
classification ($H_K$) assess the correction. At each control tick, the
governor retains the native FRR action by default and substitutes the current
cached nominal action when veto conditions are met.}
\makeatletter
\iffirstpageteaser
\IEEEaftertitletext{%
  \begin{minipage}{\textwidth}
  \centering
  \includegraphics[width=\textwidth]{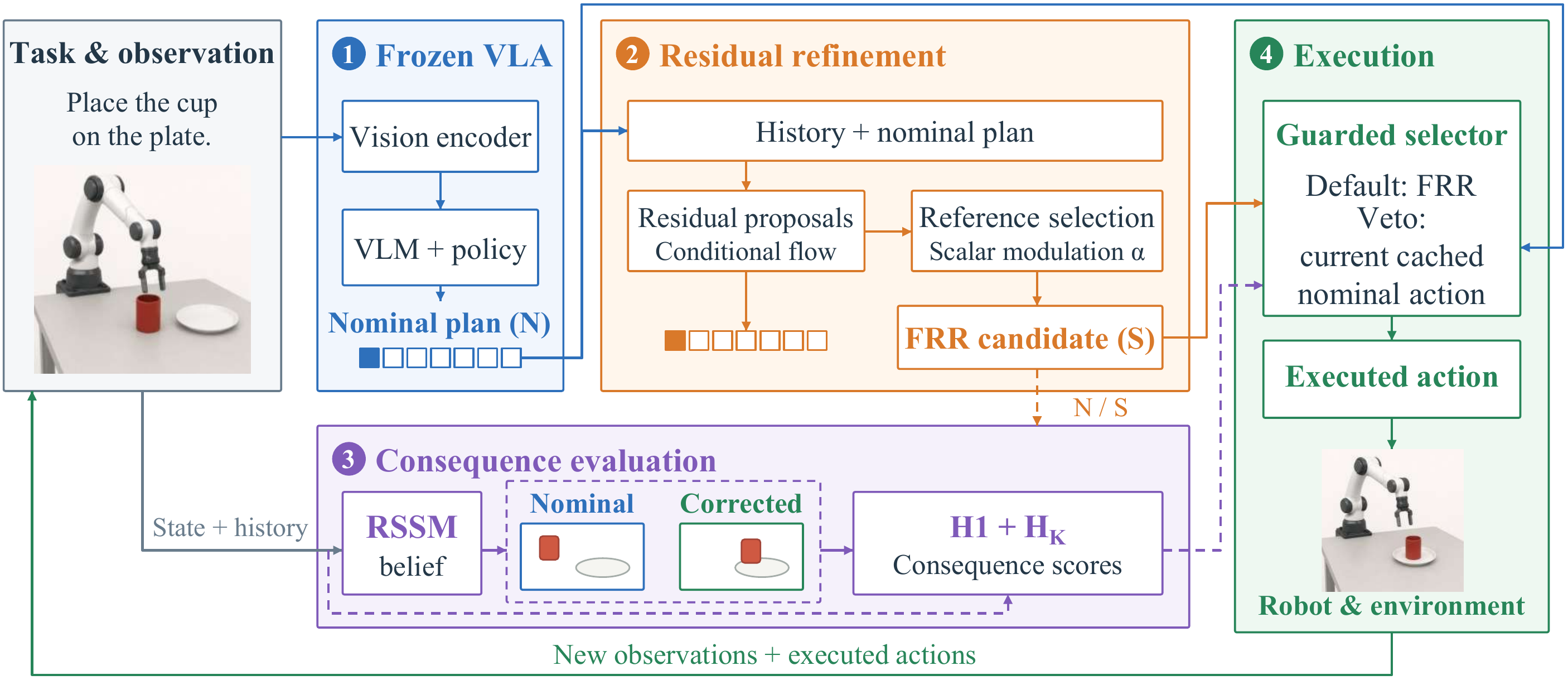}
  \def\@captype{figure}\caption{\overviewcaption}\overviewlabel
  \end{minipage}\vspace{10pt}}
\fi
\makeatother

\makeatletter
\providecommand{\bstctlcite}[1]{\@bsphack\if@filesw\immediate\write\@auxout{\string\citation{#1}}\fi\@esphack}
\makeatother
\newcommand{\toprule}{\hline}
\newcommand{\midrule}{\hline}
\newcommand{\bottomrule}{\hline}
\begin{document}
\bstctlcite{cerevla_bib_control}
\maketitle
\thispagestyle{empty}
\pagestyle{empty}
\iffirstpageteaser\else
\begin{figure*}[t]
\centering\includegraphics[width=\textwidth]{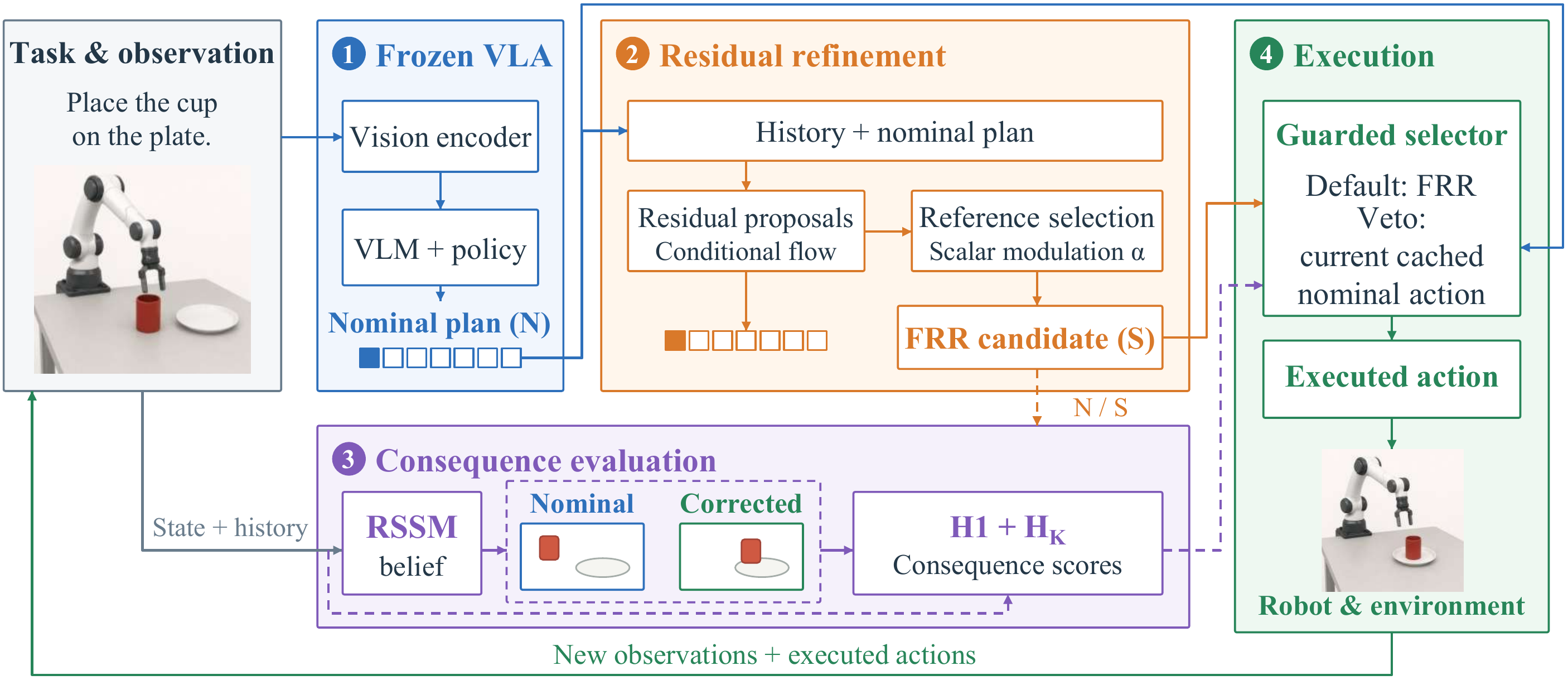}
\caption{\overviewcaption}\overviewlabel
\end{figure*}
\fi

\begin{abstract}
Action-chunked vision-language-action (VLA) policies improve
inference efficiency, but limited feedback within committed action
chunks can lead to accumulated execution errors. Residual adaptation
can correct such deviations without retraining the VLA; however,
existing corrections are typically optimized for reference-action
consistency without explicitly considering their downstream
consequences.
To address this limitation, we present Cerebellum-Inspired Consequence-Aware Residual Governance (CereVLA), a unified framework that integrates lightweight residual refinement and predictive consequence evaluation into frozen VLA execution. Corrective actions are first generated by flow-based
residual refinement, and their short- and interval-horizon
consequences are then evaluated by a recurrent state-space model and
a history-aware classifier. Residual corrections predicted to be
unfavorable are selectively suppressed by a lightweight governor.
Comparisons with state-of-the-art methods on LIBERO-10 and
LIBERO-GOAL demonstrate the effectiveness of CereVLA.
On SO-101, CereVLA increases task success from 57.5\% to 90.0\%
and reduces mean control steps by 19.6\% among successful trials,
relative to the frozen SmolVLA baseline.
\end{abstract}

\begin{figure*}[t]
\centering
\includegraphics[width=\textwidth]{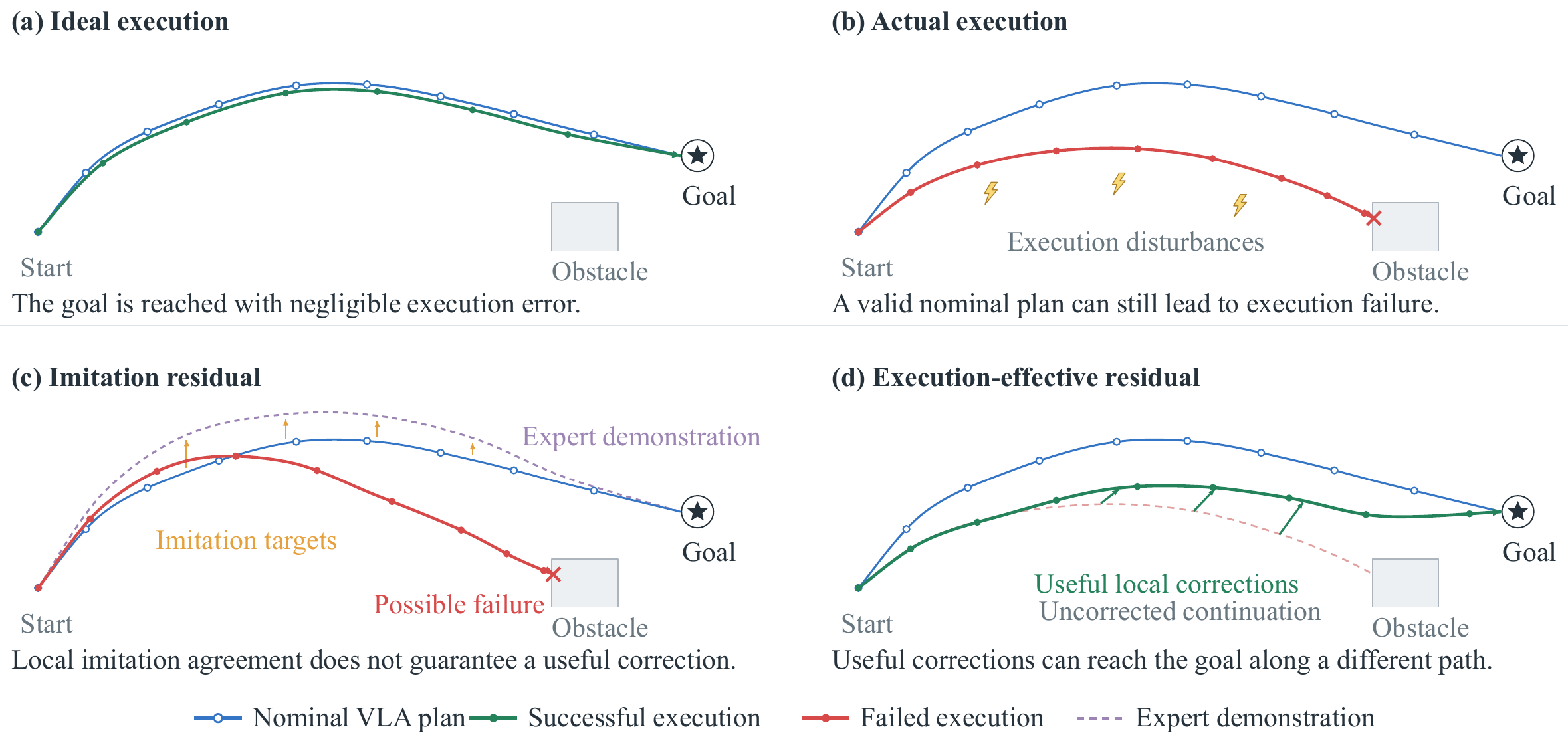}
\caption{Imitation residuals versus execution-effective corrections. (a) Ideal execution follows the nominal plan closely. (b) Disturbances can cause execution failure despite a valid nominal plan. (c) Better local agreement with an expert demonstration need not improve task completion. (d) Useful corrections may reach the goal along a different trajectory. Blue denotes nominal plans, green successful executions, red failed executions, and dashed purple an expert demonstration.}
\label{fig:residual_concept}
\end{figure*}

\section{Introduction}
\label{sec:introduction}

Vision-language-action (VLA) models are bringing robotic manipulation closer to general-purpose operation in household, industrial, laboratory, and service environments~\cite{brohan2022rt1,brohan2023rt2,kim2024openvla,octo2024,black2024pi0,physical2025pi05}. However, practical autonomy requires not only capable task reasoning, but also efficient and reliable physical execution. Repeatedly invoking a large VLA at every control step is computationally expensive, whereas executing long action chunks with limited feedback can accumulate errors as contact, tracking deviations, and environmental interactions alter the robot state.

Action chunking reduces inference cost by committing multiple actions from one policy query~\cite{zhao2023act,chi2023diffusion,kim2025oft}, but this also increases the mismatch between the observation used for planning and the state at execution time. Frequent replanning can restore feedback, yet introduces additional policy queries and does not consistently improve closed-loop behavior. Lightweight residual adaptation provides a more efficient alternative by locally correcting the nominal action without retraining the VLA. However, residual refinement is inherently double-edged: a correction may recover an execution error, but may also disturb a trajectory that would otherwise succeed.

Figure~\ref{fig:residual_concept} illustrates the key limitation. A nominal plan may fail after execution disturbances; an imitation residual may reduce local deviation from an expert demonstration yet still lead to an unfavorable continuation, whereas an execution-effective correction may depart from the demonstrated trajectory while improving task progress. Thus, the central problem is not simply how to generate a residual, but whether a generated correction should be executed. In other words, \emph{imitation agreement does not necessarily imply execution utility}. Residual generation and reactive replanning address how to modify an action or when to query the policy. CereVLA addresses the complementary decision of whether the residual-corrected action should be executed instead of the nominal action.

Inspired by the cerebellum's predictive role in estimating the consequences of motor commands~\cite{wolpert1998internal,abadia2021cerebellar}, we propose \textbf{CereVLA}, a consequence-aware execution architecture for frozen VLA policies. As shown in Fig.~\ref{fig:overview}, CereVLA separates fast residual generation from predictive residual qualification. A flow-based residual refinement (FRR) module produces lightweight online corrections from recent execution history and the cached VLA plan. CereVLA then compares nominal and corrected alternatives at complementary temporal horizons using an action-conditioned recurrent state-space model and a history-aware consequence classifier. A conservative governor suppresses a residual only when the predicted evidence indicates an adverse consequence, while otherwise preserving native residual execution and the original VLA replanning schedule. The nominal continuation serves only as a relative reference and is not assumed to be globally optimal or safe.

The main contributions of this work are summarized as follows:
\begin{itemize}
    \item We propose \textbf{CereVLA}, a unified cerebellum-inspired execution framework that integrates flow-based residual refinement, predictive consequence evaluation, and restricted-authority execution governance around frozen VLA policies.
    \item Within CereVLA, one-step consequence evaluation ($H_1$) and history-aware interval classification ($H_K$) provide complementary evidence for retaining or suppressing native FRR corrections.
    \item We evaluate CereVLA on LIBERO with two frozen VLA backbones and on a physical SO-101, reporting recovery--harm trade-offs in
simulation and task performance and motion quality on hardware.
\end{itemize}

\section{Related Work}
\label{sec:related_work}

\subsection{Action-Chunked VLA Execution}

Action chunking reduces repeated policy inference by predicting temporally extended action sequences, as adopted by ACT, Diffusion Policy, and OpenVLA-OFT~\cite{zhao2023act,chi2023diffusion,kim2025oft}. Recent methods further improve execution-time feedback. Real-Time Chunking overlaps inference with execution~\cite{black2025rtc}, while VLA-Corrector detects persistent visual deviations and adaptively truncates or replans action chunks~\cite{pan2026corrector}. These methods primarily modify when the policy is queried or replanned. CereVLA instead retains the VLA replanning schedule and addresses a complementary problem: whether an already available residual correction should be allowed to modify the committed trajectory.

\subsection{Residual Adaptation}

Residual policies augment an existing controller through additive corrections, avoiding full policy replacement or retraining~\cite{silver2018residual,johannink2018residual}. In supervised settings, residuals are naturally learned from expert--nominal action differences. Such objectives provide effective correction targets, but do not explicitly encode whether a particular residual improves downstream closed-loop execution. CereVLA therefore decouples residual proposal from residual qualification: FRR generates candidate corrections, while an independent consequence layer determines whether the native residual should be retained or suppressed.

\subsection{Predictive Consequence Modeling}

Predictive models have long been used to compensate for delayed or incomplete feedback in model-based and teleoperation control~\cite{smith2006predictor,sirouspour2006model}. Learned world models extend this idea to high-dimensional robotic decision making; Dreamer learns latent dynamics for behavior optimization~\cite{hafner2019dream}, while Generative Predictive Control evaluates or refines proposals from frozen generative policies~\cite{qi2026gpc}. VIP and LIV provide complementary visual representations of task progress~\cite{ma2022vip,ma2023liv}. Unlike methods that perform full trajectory imagination or policy replanning, CereVLA uses predictive modeling only to compare the relative consequences of nominal and residual actions. This narrower role enables lightweight execution-time governance without modifying the frozen VLA.

\begin{figure*}[t]
\centering\includegraphics[width=\textwidth]{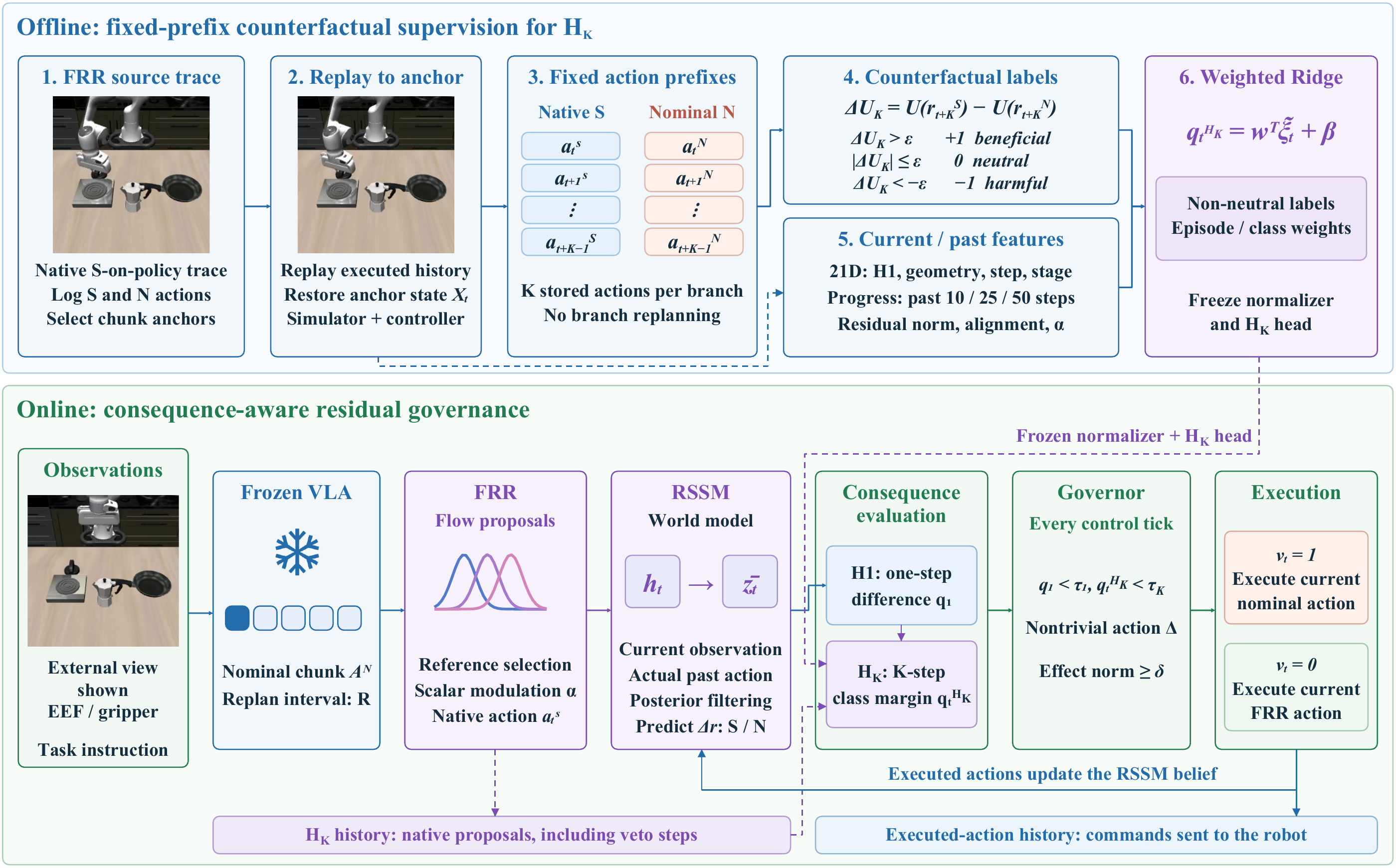}
\caption{Offline supervision and online execution in CereVLA. Recorded native and
nominal $K$-action prefixes are replayed from the same anchor without branch
replanning. Endpoint utility differences provide signed labels for $H_K$.
Online inference uses current and past features with the frozen classifier
and its fitted normalizer; recorded future actions are used only for offline
labeling. The governor recomputes its decision at every control tick. Each
veto substitutes the current cached nominal action only and may recur without
an episode-wide count limit. $R$ is the VLA replanning interval; $H_5$ is the
evaluated $K=5$ instance.}
\label{fig:pathways}
\end{figure*}

\section{Methodology}
\label{sec:method}

CereVLA evaluates residual corrections before they modify the execution of a
frozen vision-language-action (VLA) policy. The framework combines flow-based
residual refinement (FRR), one-step consequence evaluation ($H_1$),
interval-level consequence classification ($H_K$), and an execution governor.
We distinguish nominal VLA execution $N$, native FRR execution $S$, and
governed execution $G$. The VLA replanning interval is denoted by $R$, whereas
$K$ denotes the offline consequence-supervision horizon of $H_K$. The reported
implementation uses $R=K=5$, and the corresponding interval classifier is
denoted by $H_5$. Importantly, the online governor is evaluated at every control
tick; $K$ does not specify the duration of an online fallback.

\subsection{Frozen Planning and Residual Refinement}
\label{subsec:problem}

The VLA generates a nominal action chunk every $R$ control steps.
Between replanning events, FRR uses the latest feedback to refine the current
action. Its input is
\begin{equation}
c_t^{\mathrm{FRR}}=
\bigl(s_{t-7:t},\bar{\mathbf A}_t^N,a_t^N,
e_{\mathrm{task}},\rho_t,z_t^{\mathrm{vis}}\bigr),
\label{eq:frr_context}
\end{equation}
where $s_t\in\mathbb R^8$ contains end-effector (EEF) position, an
axis-angle orientation vector, and two gripper coordinates. The nominal plan
$\bar{\mathbf A}_t^N\in\mathbb R^{50\times7}$ is shifted to the current
step and zero-padded, with first action $a_t^N$. The task code
$e_{\mathrm{task}}$ is a one-hot task identifier, and $\rho_t$ is plan age
divided by $R-1$. Visual context $z_t^{\mathrm{vis}}\in\mathbb R^{1920}$
concatenates mean-pooled frozen-image-encoder features from two camera views.

FRR learns aligned expert--nominal residuals through conditional flow
matching~\cite{lipman2023flow}:
\begin{align}
\mathbf R^*&=\operatorname{clip}\!\left(
(\mathbf A^E-\bar{\mathbf A}^N)\oslash\boldsymbol s_\Delta,
-1,1\right),\nonumber\\
\mathbf R_u&=(1-u)\epsilon+u\mathbf R^*,\nonumber\\
\mathcal L_{\mathrm{flow}}
&=\mathbb E\!\left[
\operatorname{MSE}_{M}\!\left(
v_\theta(\mathbf R_u,u,c_t^{\mathrm{FRR}},m_t),
\mathbf R^*-\epsilon\right)\right].
\label{eq:flow_loss}
\end{align}
Here $\mathbf A^E$ denotes the aligned expert action sequence,
$\epsilon\sim\mathcal N(0,I)$, $u\sim\mathcal U(0,1)$, and
$\boldsymbol s_\Delta$ contains the stored per-action residual scales. The
mask $M$ combines chunk validity and correction supervision. The binary
generation mode $m_t$ uses correction labels during training and is determined
by auxiliary risk/recovery heads during inference. Auxiliary objectives
supervise the expert-reference and local-dynamics heads, together with risk,
progress, recovery, replanning, and value estimates; flow-velocity
regularization is also applied.

Four residual candidates are sampled using four Heun integration steps. Each
normalized sample $\widehat{\mathbf R}_t^{(i)}$ is converted to an
action-space residual,
\begin{equation}
\Delta\mathbf A_t^{(i)}=
\operatorname{clip}(\widehat{\mathbf R}_t^{(i)},-1,1)
\odot\boldsymbol s_\Delta.
\label{eq:frr_residual_scale}
\end{equation}
Each corrected plan $\bar{\mathbf A}_t^N+\Delta\mathbf A_t^{(i)}$ is
clipped to the action bounds before reference-based selection. The candidate
set also includes the unmodified nominal plan with zero residual. An
expert-supervised reference head selects the candidate with the closest first
action:
\begin{equation}
\begin{aligned}
a_t^{\mathrm{ref}}&=\tanh f_{\mathrm{ref}}(c_t^{\mathrm{FRR}}),\\
i^*&=\arg\min_i\|a_{t,0}^{(i)}-a_t^{\mathrm{ref}}\|_2^2.
\end{aligned}
\label{eq:frr_selection}
\end{equation}
Let $\Delta a_{t,0}^{(i)}$ denote the first action-space residual of
candidate $i$. The native command is
\begin{equation}
a_t^S=a_t^N+\alpha_t\Delta a_{t,0}^{(i^*)},
\qquad
\alpha_t=\operatorname{sigmoid}(f_\alpha(c_t^{\mathrm{FRR}})).
\label{eq:native_action}
\end{equation}
Only the current command is used for execution; residual proposals are
recomputed at the next control step from updated feedback. If the native output
is non-finite, the controller executes $a_t^N$ instead. CereVLA preserves this
native FRR computation and evaluates the effective correction
$\Delta a_t=a_t^S-a_t^N$ before execution.

\subsection{Task Relations and One-Step Consequences}
\label{subsec:predictive}
\label{subsec:h1}

For task $k$, the relation map $r_t=\Phi_k(X_t)$ encodes object--EEF and
object--goal offsets, together with articulated-joint error when applicable.
Here $X_t$ denotes the underlying robot--environment state. The current task
stage determines which relation is relevant to progress.

A recurrent state-space model (RSSM)~\cite{hafner2018planet,hafner2025dreamer}
estimates action-dependent changes in these relations. Its input $x_t$ includes
robot state, the previous executed action, the current nominal action, a
five-step nominal-plan window, plan phase and age, task code, and visual
context. With $e_t=E_\psi(x_t)$, online filtering uses the posterior mean:
\begin{equation}
\bar z_t=\mu_{q,\psi}(h_t,e_t),
\qquad
h_{t+1}=\operatorname{GRU}_\psi
([e_t,\bar z_t,a_t^{\mathrm{exec}}],h_t).
\label{eq:belief}
\end{equation}
The RSSM is trained with stochastic latents using reconstruction, robot- and
relation-change prediction, latent KL, and auxiliary predicate objectives.
Predicted relation changes provide the consequences used for action comparison.

At every control tick with a nontrivial native correction, nominal and native
FRR actions are evaluated from the same filtered belief and observation
encoding. Four shared prior-noise samples produce paired predictions, which are
transformed back to relation units and averaged. Holding the current task stage
and target fixed, $H_1$ computes
\begin{align}
\hat r_{t+1}^B&=r_t+\Delta\hat r_t^B,
\quad B\in\{N,S\},\nonumber\\
q_t^{H_1}&=U_t(\hat r_{t+1}^S)-U_t(\hat r_{t+1}^N).
\label{eq:h1_score}
\end{align}
The utility $U_t$ is the negative object--EEF distance during approach,
negative object--goal distance during placement, and negative absolute joint
error during closure. Thus, $q_t^{H_1}<0$ indicates a predicted adverse
one-step consequence of the native correction.

The paired formulation emphasizes \emph{relative} consequence prediction.
Let $e_t^S$ and $e_t^N$ denote the utility-prediction errors of the native
and nominal branches, and let $\Delta U_{1,t}$ be their true one-step utility
difference. Then
\begin{equation}
q_t^{H_1}-\Delta U_{1,t}
=
e_t^S-e_t^N.
\label{eq:paired_error}
\end{equation}
Hence the comparison error depends on the difference between branch-wise
prediction errors rather than on either absolute prediction error alone.
Sharing the anchor, belief, predictor, and stochastic samples encourages
common-mode errors to cancel, although such cancellation is not guaranteed.

\subsection{Interval-Level Counterfactual Supervision}
\label{subsec:h5}

$H_K$ complements the one-step comparison by classifying consequences over a
fixed $K$-action horizon. Source trajectories are collected under frozen native
FRR execution. For each selected anchor, two branches replay the same recorded
execution history and then apply either the recorded native or nominal prefix:
\begin{equation}
\begin{aligned}
r^B_{t+K}
&=\Phi_k\!\left(
\mathcal{T}_K(X_t,\mathbf A^B_{t,\mathrm{src}})
\right),
\quad B\in\{N,S\},\\
\Delta U_{K,t}
&=U_t(r^S_{t+K})-U_t(r^N_{t+K}).
\end{aligned}
\label{eq:h5_oracle}
\end{equation}
Here $\mathcal{T}_K$ denotes evolution under $K$ prescribed actions, with
$\mathbf A^B_{t,\mathrm{src}}$ containing the source actions from $t$
through $t+K-1$. Both branches keep the recorded prefixes fixed, so the
labels characterize these prefixes rather than feedback policies recomputed
along each branch. The target is $y_t=+1$ when
$\Delta U_{K,t}>\epsilon$, $y_t=-1$ when
$\Delta U_{K,t}<-\epsilon$, and $y_t=0$ otherwise. Neutral records are
retained but excluded from classifier fitting and classwise decision metrics.

In the implemented five-step instance, the feature vector
$\xi_t\in\mathbb R^{21}$ contains the $H_1$ score, the log-magnitude of the
predicted relation difference, current task distances and closure error,
normalized episode step, stage progress over 10/25/50-step windows, window
means of residual norm, nominal--native cosine alignment and $\alpha_t$, and
the task-stage one-hot code. Motion-history features use native FRR proposals,
including during vetoed intervals; the RSSM instead updates its belief using
the actions actually executed.

For standardized features $\tilde\xi_i$, $H_K$ is fitted using a weighted
ridge classification surrogate:
\begin{align}
(w^*,\beta^*)&=\arg\min_{w,\beta}
\sum_{i:y_i\neq0}
\omega_i(y_i-w^\top\tilde\xi_i-\beta)^2\nonumber\\
&\hspace{10mm}+\lambda\|w\|_2^2+10^{-8}\beta^2,\nonumber\\
q_t^{H_K}&=(w^*)^\top\tilde\xi_t+\beta^*.
\label{eq:h5_ridge}
\end{align}
Sample weights combine inverse per-episode counts with class-frequency
weighting and are normalized to unit mean. Lower scores indicate stronger
evidence of a harmful prefix. Unlike $H_1$, $H_K$ outputs a signed
classification margin rather than a predicted utility difference. At
deployment, it evaluates current and past features without an online
$K$-step world-model rollout.

\noindent\textbf{Counterfactual decision interpretation.}
The offline supervision in Eq.~\eqref{eq:h5_oracle} can be interpreted as a
selection problem between native and nominal fixed prefixes. For an idealized
selector, let $v\in\{0,1\}$ denote selecting the nominal prefix
($v=1$) rather than the native FRR prefix ($v=0$). Relative to an oracle that
always selects the higher-utility branch, the local selection regret is
\begin{equation}
\begin{aligned}
\mathcal R_t(v)
&=v[\Delta U_{K,t}]_+
 +(1-v)[-\Delta U_{K,t}]_+,\\
[x]_+&\triangleq \max(x,0).
\end{aligned}
\label{eq:governance_regret}
\end{equation}
Thus, an incorrect rejection of a strongly beneficial residual can incur a
larger consequence loss than failing to reject a mildly harmful residual.
This perspective motivates consequence-aware qualification based on both sign
and consequence magnitude rather than classification accuracy alone.

Equation~\eqref{eq:governance_regret} characterizes only the fixed-prefix
counterfactual supervision used to train $H_K$. It is not an equality for the
deployed CereVLA trajectory: online execution re-evaluates the decision at
every control tick and does not commit to either fixed $K$-step branch.

\subsection{Execution Governance}
\label{subsec:governor}

At every control tick, CereVLA combines the short-horizon $H_1$ evidence and
the interval-level $H_K$ margin to determine whether the current native FRR
correction should be retained. Let $d_t\in\{0,1\}$ indicate that the native
and nominal actions differ by a nontrivial amount, and define the predicted
relative relation change
\begin{equation}
\Delta\hat r_t^{\mathrm{CF}}
=
\Delta\hat r_t^S-\Delta\hat r_t^N.
\label{eq:cf_relation}
\end{equation}
The veto rule is
\begin{equation}
v_t
=
d_t\,
\mathbf 1[q_t^{H_1}<\tau_1]\,
\mathbf 1[q_t^{H_K}<\tau_K]\,
\mathbf 1[
\|\Delta\hat r_t^{\mathrm{CF}}\|_2\ge\delta
].
\label{eq:veto}
\end{equation}
When $v_t=0$, CereVLA retains the native FRR command. When $v_t=1$, only
the current residual correction is suppressed and the corresponding nominal
action is executed:
\begin{equation}
a_t^G
=
(1-v_t)a_t^S+v_ta_t^N.
\label{eq:governed_action}
\end{equation}

The decision is recomputed at the next control tick using the updated
observation, FRR proposal, and executed-action history. Consequently, vetoes
are repeatable and there is no episode-level intervention budget. Each veto
acts only on the current action and does not lock execution to a nominal
$K$-step prefix. The horizon $K$ therefore defines the interval used to
construct supervision for $H_K$, whereas online governance remains
temporally local and feedback-driven.

This design gives the consequence layer restricted authority: it may suppress
the current residual and revert to the nominal action, but it cannot generate
an arbitrary alternative command or alter the VLA replanning schedule. The
nominal action is therefore used as a relative execution reference rather than
as a certified safe controller.

\section{Experiments}
\label{sec:experiments}

\subsection{Experimental Setup}
\label{subsec:exp_protocol}

We evaluate frozen $\pi_{0.5}$ and SmolVLA policies on LIBERO-10
and LIBERO-GOAL~\cite{liu2023libero}. Episodes terminate at success
or the task horizon; failures and timeouts remain in the success-rate
denominator. The standard policy replans every $R=5$ control steps,
while FRR updates the current correction at each step. The evaluated interval classifier is $H_5$, with $K=5$. Paired
comparisons use matched task initial conditions. The current
workstation contains an NVIDIA RTX 4090 GPU and an Intel Core
i7-13700 CPU.

We compare the nominal \emph{Base} policy, \emph{Frequent replanning}
at every control step, native \emph{FRR}, its \emph{modulation variant},
and \emph{CereVLA}. VLA-Corrector~\cite{pan2026corrector} and FutureRTC~\cite{jiang2026futurertc}
provide additional execution-correction comparisons under their
evaluated configurations. FRR tests residual adaptation, whereas
CereVLA additionally governs whether the proposed correction is used.

We report suite success rates and their equal-weight average.
Relative to the corresponding Base execution, Recovery ($R$) and
Harm ($H$) are the percentages of all paired episodes that change
from failure to success and success to failure, respectively.
If $n_{ij}$ counts reference/method outcomes $(i,j)$ and
$n=\sum_{i,j}n_{ij}$, then $R=100n_{01}/n$, $H=100n_{10}/n$, and
$\mathrm{Net}=R-H$. Net is the paired success difference in
percentage points.

\begin{table*}[!t]
\centering
\footnotesize
\setlength{\tabcolsep}{3.5pt}
\renewcommand{\arraystretch}{1.05}
\caption{Task success and paired recovery--harm analysis on LIBERO.
Success, $R$, and $H$ are percentages; Net is in percentage points.
Avg. is the equal-suite mean; paired outcomes use Base as reference.}
\label{tab:libero_main}
\begin{tabular}{@{}llrrrrr@{}}
\toprule
Backbone & Method & LIBERO-10 & LIBERO-GOAL & Avg.
& \shortstack{L10\\R/H/Net} & \shortstack{GOAL\\R/H/Net}\\
\midrule
$\pi_{0.5}$ & Base
& 92.00 & 96.00 & 94.00 & -- & --\\
& Frequent replanning
& 69.50 & 89.50 & 79.50
& $3.50/26.00/-22.50$ & $2.00/8.50/-6.50$\\
& FRR
& 93.00 & 97.00 & 95.00
& $6.50/5.50/+1.00$ & $2.00/1.00/+1.00$\\
& FRR (modulation variant)
& 93.00 & 97.50 & 95.25
& $4.50/3.50/+1.00$ & $2.50/1.00/+1.50$\\
& VLA-Corrector
& 92.00 & 96.50 & 94.25
& $3.50/3.50/0.00$ & $2.50/2.00/+0.50$\\
& FutureRTC
& 91.50 & \textbf{98.50} & 95.00
& $2.00/2.50/-0.50$ & $3.50/1.00/+2.50$\\
& CereVLA
& \textbf{93.20} & 97.50 & \textbf{95.35}
& $4.40/3.20/+1.20$ & $3.25/1.75/+1.50$\\
\midrule
SmolVLA & Base
& 46.25 & 68.00 & 57.13 & -- & --\\
& Frequent replanning
& 27.50 & \textbf{79.50} & 53.50
& $3.25/22.00/-18.75$ & $22.00/10.50/+11.50$\\
& FRR
& 46.50 & 72.00 & 59.25
& $13.25/13.00/+0.25$ & $11.00/7.00/+4.00$\\
& FRR (modulation variant)
& 47.00 & 72.00 & 59.50
& $6.25/5.50/+0.75$ & $11.00/7.00/+4.00$\\
& VLA-Corrector
& 20.50 & 50.50 & 35.50
& $10.50/36.25/-25.75$ & $22.50/40.00/-17.50$\\
& FutureRTC
& 20.00 & 49.50 & 34.75
& $13.00/39.25/-26.25$ & $15.00/33.50/-18.50$\\
& CereVLA
& \textbf{55.00} & 73.00 & \textbf{64.00}
& $15.50/6.75/+8.75$ & $9.50/4.50/+5.00$\\
\bottomrule
\end{tabular}
\par\smallskip
\begin{minipage}{\textwidth}\footnotesize
Bold marks the highest success rate per column within each backbone.
\end{minipage}
\end{table*}

\subsection{Overall Performance}
\label{subsec:exp_main}

Table~\ref{tab:libero_main} shows that CereVLA has the highest
reported equal-suite mean for each backbone, reaching 95.35\% with
$\pi_{0.5}$ and 64.00\% with SmolVLA. Both improve over Base and
native FRR on the two suites. The advantage is more pronounced
with SmolVLA; for $\pi_{0.5}$, the mean gain over FRR is modest,
and FutureRTC retains the highest LIBERO-GOAL success rate.
The results therefore favor CereVLA in aggregate without implying
superiority on every backbone--suite combination.

The paired Recovery--Harm analysis provides a more direct view of execution reliability than aggregate success alone. Native FRR exhibits a clear recovery--harm trade-off: residual corrections can recover failed nominal executions, but the same mechanism can also convert previously successful executions into failures. CereVLA shifts this balance toward more selective intervention. Across all four backbone--suite combinations, CereVLA achieves a positive Net gain that exceeds native FRR, indicating that its improvement is not driven simply by applying more corrections, but by producing a more favorable balance between recovered failures and newly introduced failures.

This effect is particularly evident with SmolVLA. On LIBERO-10, CereVLA substantially reduces Harm relative to native FRR while simultaneously increasing Recovery, yielding a markedly larger positive Net gain. On LIBERO-GOAL, CereVLA again reduces Harm while maintaining sufficient Recovery to improve the paired Net outcome. For $\pi_{0.5}$, the gain is more moderate and does not arise from uniformly minimizing Harm. Instead, CereVLA maintains a favorable recovery--harm balance, producing a higher positive Net outcome on both suites. This distinction is important: the governor is designed to improve closed-loop utility rather than to suppress residual activity as aggressively as possible. These results support the intended role of consequence-aware governance: beneficial residual behavior is largely retained, whereas corrections with adverse execution consequences are more selectively suppressed.

\subsection{Component Analysis and Motion Quality}
\label{subsec:exp_ablation}
\label{subsec:exp_evaluator}
\label{subsec:exp_motion}

We use LIBERO to examine local consequence discrimination and
execution quality with SmolVLA + FRR. Beneficial retention (BR)
is the fraction of beneficial labeled anchors retained; harmful
rejection (HR) is the fraction of harmful anchors vetoed. Veto
precision is the harmful fraction among vetoed non-neutral anchors.
These local metrics precede execution governance and differ from
episode-level Recovery and Harm.

\begin{table*}[t]
\centering
\footnotesize
\setlength{\tabcolsep}{4pt}
\renewcommand{\arraystretch}{1.15}

\caption{
Evaluator component analysis on LIBERO with SmolVLA.
Local BR/HR metrics evaluate consequence-based decisions,
while motion metrics and evaluator overhead are measured
relative to native FRR execution.
}

\label{tab:mechanism_efficiency}

\begin{tabular}{lrrrrrrrr}
\toprule

Evaluator
&
\shortstack{BR\\(\%)}
&
\shortstack{HR\\(\%)}
&
\shortstack{Veto\\Prec.(\%)}
&
\shortstack{Steps\\$\Delta$(\%)}
&
\shortstack{Action\\D3 $\Delta$(\%)}
&
\shortstack{EEF Pos.\\Jerk $\Delta$(\%)}
&
\shortstack{Rot. Vec.\\D3 $\Delta$(\%)}
&
\shortstack{Eval.\\Overhead(ms)}
\\

\midrule

H1 geometric
&
100.0
&
100.0
&
100.0
&
+38.67
&
-18.30
&
-4.10
&
-16.26
&
0.002
\\

H1 learned
&
75.0
&
80.0
&
80.0
&
+18.29
&
-30.45
&
-6.76
&
-10.40
&
2.006
\\

H1 + H5
&
100.0
&
53.3
&
100.0
&
+9.37
&
-16.61
&
+2.30
&
+7.22
&
2.733
\\

+ Continuation
&
100.0
&
53.3
&
100.0
&
-2.34
&
+18.21
&
+19.72
&
+19.57
&
2.737
\\

Full (+ Governor)
&
100.0
&
53.3
&
100.0
&
-9.35
&
-3.68
&
-1.71
&
-4.92
&
2.740
\\

\bottomrule

\end{tabular}

\end{table*}

\begin{table*}[t]
\centering
\small
\setlength{\tabcolsep}{5pt}
\caption{Real-robot performance on the SO-101 orange pick-and-place task.
Each method uses 40 trials. Steps and motion metrics are reported over
successful trials as mean $\pm$ sample standard deviation.}
\label{tab:real_robot}
\begin{tabular}{lccccc}
\toprule
Method &
\shortstack{Success} &
\shortstack{Steps} &
\shortstack{Span (s)} &
\shortstack{Command Jerk\\Proxy } &
\shortstack{Acceleration RMS\\(rad/s$^2$) } \\
\midrule
SmolVLA
& 57.5\% 
& $354.70\pm89.78$
& $12.71\pm3.23$
& $\mathbf{3.462\pm0.446}$
& $4.383\pm0.465$ \\

SmolVLA + FRR
& 82.5\% 
& $322.21\pm65.55$
& $11.66\pm2.30$
& $4.319\pm0.406$
& $4.588\pm0.496$ \\

\textbf{CereVLA}
& \textbf{90.0\% }
& $\mathbf{285.34\pm30.45}$
& $\mathbf{9.97\pm1.06}$
& $4.020\pm0.333$
& $\mathbf{4.329\pm0.498}$ \\
\bottomrule
\end{tabular}
\end{table*}

Table~\ref{tab:mechanism_efficiency} shows that adding H5 shifts the evaluator toward more conservative intervention: beneficial retention and veto precision improve, while harmful rejection decreases. This trade-off is intentional, since CereVLA prioritizes preserving useful residuals rather than maximizing veto frequency.

More importantly, H1+H5 and the Full Governor share the same local BR, HR, and veto-precision scores, yet produce markedly different closed-loop behavior. The Full Governor is the only configuration that simultaneously reduces completion steps, action variation, and both translational and rotational EEF variation. This indicates that consequence prediction alone is insufficient; bounded governance is required to translate local evidence into efficient and smooth execution.

Motion metrics use third finite differences of the logged executed actions and EEF trajectories. Action D3 RMS is computed over all seven command dimensions without time scaling. EEF position and axis-angle rotation-vector differences are divided by $\Delta t^3$ with $\Delta t=0.05$\,s, where the orientation measure is treated as a coordinate-based smoothness proxy. Metrics are averaged over the common-success subset, and relative changes are computed from these episode-level means.

\paragraph{Computational overhead}
The CereVLA execution layer introduces 8.06M additional parameters. This
corresponds to 1.33\% of the 604.93M-parameter frozen SmolVLA backbone
and approximately 0.27\% of a 3B-parameter $\pi_{0.5}$ backbone. On the
RTX 4090 workstation, FRR inference requires 8.2\,ms, while the
consequence module adds 2.74\,ms per tick. These timings exclude VLA
inference and therefore should not be interpreted as total control-loop
latency.

\subsection{Real-Robot Evaluation}
\label{subsec:real_robot}

\noindent\textbf{Setup and metrics.}
We evaluate CereVLA on a physical SO-101 using the instruction
``Grab the orange to basket.'' Before each trial, the orange is randomly
placed within a predefined reachable tabletop workspace, while the target
basket remains fixed. The robot uses a front-view camera, a wrist camera,
and joint-state feedback. All methods share the same frozen SmolVLA
checkpoint.

We compare \emph{Base}, native \emph{FRR}, and the complete
\emph{CereVLA}. FRR recomputes the current residual from updated feedback
at every control cycle, while SmolVLA retains its native 50-action queue.
CereVLA further evaluates nominal and residual-corrected alternatives using
short- and interval-horizon consequence evidence and permits repeated vetoes
without an episode-level budget. Each method is evaluated over 40 trials.
The target control rate is 30\,Hz, with a 20\,s / 700-step task limit.
Success requires placing the orange inside the basket. The three groups are
collected in separate batches and are therefore not strictly paired.

\begin{figure}[t]
    \centering
    \includegraphics[width=\columnwidth]{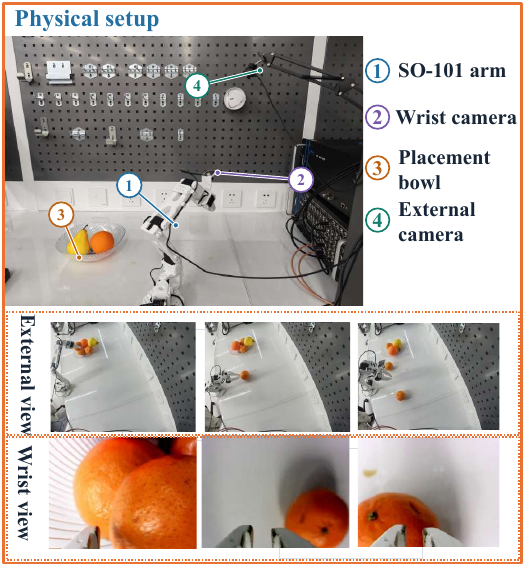}
    \caption{Physical SO-101 setup for orange pick-and-place with randomized
    object placement, wrist camera, and external camera.}
    \label{fig:real_setup}
\end{figure}

Task efficiency is measured by successful-trial steps and recorded execution
span. Motion quality is evaluated using the RMS third finite difference of
the five arm-joint commands (command jerk proxy) and the time-weighted RMS
of measured joint-acceleration magnitude. Joint velocity and acceleration
are computed offline from measured joint positions and physical timestamps.

\noindent\textbf{Results.}
Native FRR substantially improves real-robot success over the frozen
SmolVLA baseline, but also increases command variation. CereVLA further
raises success to 90.0\%, while reducing mean completion steps and execution
span by 11.4\% and 14.5\%, respectively, relative to FRR. It also reduces
the FRR-induced command jerk by 6.9\% and measured acceleration RMS by
5.6\%. Thus, CereVLA preserves the recovery capability of residual
refinement while improving execution efficiency and mitigating part of its
additional motion variation.

\begin{figure}[t]
    \centering
    \includegraphics[width=\columnwidth]{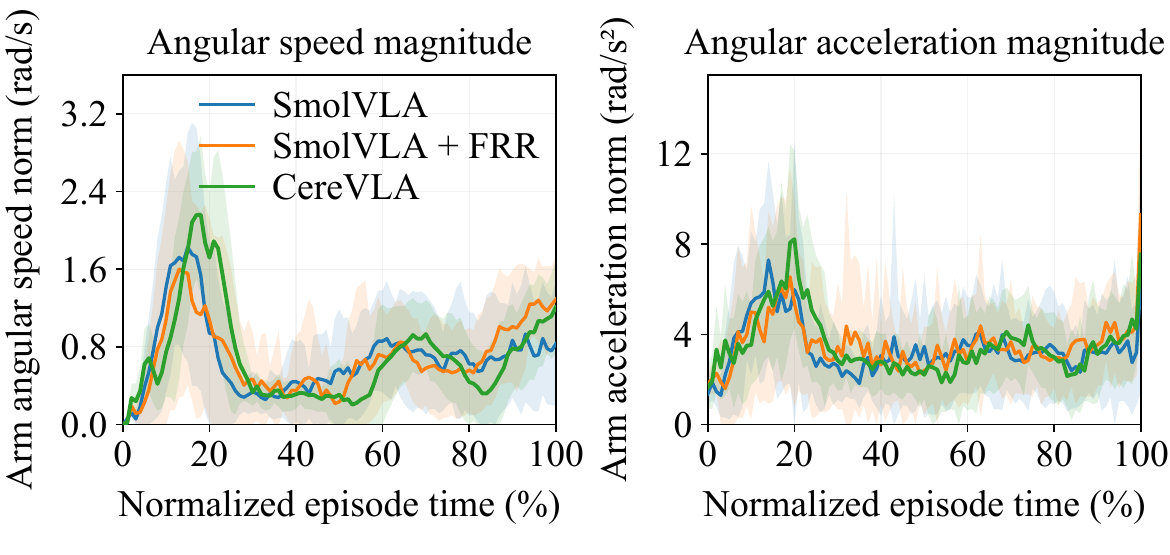}
    \caption{Measured whole-arm angular-speed and angular-acceleration
    magnitudes over normalized successful-episode time. Curves show the
    across-episode mean and shaded regions indicate one standard deviation.}
    \label{fig:real_motion}
\end{figure}

Figure~\ref{fig:real_motion} provides the corresponding time-resolved motion
comparison. The nominal policy retains the lowest command jerk, whereas
CereVLA reduces the additional variation introduced by FRR while achieving
higher task success. Successful CereVLA runs operate at an effective
observation rate of approximately 28.5\,Hz under the configured 30\,Hz
target. Since the three groups are not paired by initial state, these results
are interpreted as deployment-level evidence rather than a strictly paired
causal comparison.

\section{Conclusion}
\label{sec:conclusion}

This work presents \textbf{CereVLA}, a consequence-aware execution
framework that combines online residual refinement with predictive
qualification for frozen action-chunked VLA policies. Across LIBERO-10
and LIBERO-GOAL, CereVLA improves task success on two VLA backbones and
achieves a more favorable Recovery--Harm balance than native residual
execution. Real-robot experiments on a low-cost SO-101 further increase
success from 82.5\% with FRR to 90.0\%, while reducing completion steps
and part of the residual-induced motion variation. These results show that
reliable VLA execution depends not only on generating corrective residuals,
but also on predicting when those corrections should be applied. Future
work will explore longer-horizon world-model reasoning and
reinforcement-learning signals for optimizing residual generation toward
closed-loop task utility.

\balance
\bibliographystyle{IEEEtran}
\bibliography{bibstyle_controls,references}
\end{document}